\documentclass{article}

 \usepackage[nonatbib, final]{nips_2017}

\usepackage{lineno}

\usepackage[utf8]{inputenc} 
\usepackage[T1]{fontenc}    
\usepackage{hyperref}       
\usepackage{url}            
\usepackage{booktabs}       
\usepackage{amsfonts}       
\usepackage{nicefrac}       
\usepackage{microtype}      

\usepackage{times}
\usepackage{latexsym}

\usepackage{graphicx}
\usepackage{bbm}
\def\R{{\mathbbm R}}
\usepackage{array}
\newcolumntype{P}[1]{>{\centering\arraybackslash}p{#1}}
\usepackage{subfigure}
\usepackage{epstopdf}
\usepackage{mathtools}
\usepackage{xcolor}
\usepackage{wrapfig, lipsum, booktabs, multicol, multirow}
\usepackage{adjustbox}
\newcommand*{\affaddr}[1]{#1}
\newcommand*{\affmark}[1][*]{\textsuperscript{#1}}

\title{Multi-task Memory Networks for Category-specific Aspect and Opinion Terms Co-extraction}

  \author{Wenya Wang\affmark[\dag\ddag]
  	\hspace{3mm}
  	Sinno Jialin Pan\affmark[\dag]
  	\hspace{3mm}
  	Daniel Dahlmeier\affmark[\ddag]
  	\\
  	\affaddr{\affmark[\dag]Nanyang Technological University, Singapore} \\
  	\affaddr{\affmark[\ddag]SAP Innovation Center Singapore} \\
  	\affmark[\dag]\{wa0001ya, sinnopan\}@ntu.edu.sg, \affmark[\ddag]\{d.dahlmeier\}@sap.com
  }

\begin{document}
\maketitle
\begin{abstract}
In aspect-based sentiment analysis, most existing methods focus on either identifying aspect/opinion terms or categorizing pre-extracted aspect terms. Each task by itself only provides partial information to end users. To generate more detailed and structured opinion analysis, we study a finer-grained problem, which we call category-specific aspect and opinion terms extraction. This problem involves identification of aspect and opinion terms within each sentence, as well as categorization of the identified terms at the same time. To this end, we propose an end-to-end multi-task memory network, where aspect/opinion terms extraction for a specific category is considered as a task, and all the tasks are learned jointly by exploring commonalities and relationships among them. We demonstrate state-of-the-art performance of our proposed model on three benchmark datasets.
\end{abstract}

\section{Introduction}

Aspect-based sentiment analysis aims to provide fine-grained information via token-level predictions. Under this branch, a number of work has been proposed for aspect/opinion terms extraction~\cite{Hu04,Qiu11,Wang16}. Here, an aspect term refers to a word or a phrase describing some feature of an entity, and an opinion term refers to the expression carrying subjective emotions. For example, in the sentence ``\textit{The soup is served with nice portion, the service is prompt}'', \textit{soup}, \textit{portion} and \textit{service} are aspect terms, while \textit{nice} and \textit{prompt} are opinion terms. This extraction task simply identifies aspect/opinion terms without classifying them into different categories, e.g., \textit{soup} is under the ``DRINK'' category, while \textit{service} is under the ``SERVICE'' category. On the other hand, some previous works focus on categorization of aspect terms, where aspect terms are extracted in advance, and the goal is to classify it into one of the predefined categories~\cite{Carenini05,Yu2011,Zhai10,Zhai11}. Although topic models~\cite{Guo09,Titov08} can achieve both grouping and extraction at the same time, they mainly focus on grouping, and could only identify general and coarse-grained aspect terms.

To provide a more useful end-to-end analysis, we introduce a finer-grained task named \textit{category-specific aspect and opinion terms extraction}, where aspect/opinion terms need to be extracted and classified to a category from a pre-defined set, simultaneously. This is also benefitial for linking aspect terms and opinion terms through their category information. Consider the previous example, our objective is to extract and classify \textit{soup} and \textit{portion} as aspect terms under the ``DRINKS'' category, and \textit{service} as an aspect term under the ``SERVICE'' category, similar for the opinion terms {\em nice} and {\em prompt}. The proposed task is much more challenging because when specific categories are taken into consideration for terms extraction, training data may become extremely sparse, e.g. certain categories may only contain very few reviews or sentences. Moreover, it requires to achieve both extraction and categorization, simultaneously, which significantly increases the difficulty compared with the task of only extracting overall aspect/opinion terms or classifying pre-extracted terms.

In this paper, we offer an end-to-end deep multi-task learning architecture to accomplish the proposed task. Our high-level idea is that we consider terms extraction for each specific category as an individual task, and design a memory network to co-extract aspect terms and opinion terms within each task via dual label propagation. The memory networks are then jointly learned in a multi-task learning network to address the data sparsity issue of each task. We conduct extensive experiments on SemEval Challenge benchmark datasets to demonstrate state-of-the-art performance of our proposed approach.


\section{Related Work}
There have been a number of works proposed for aspect/opinion terms extraction~\cite{Hu04,Pop05,Zhuang06,Wu09,Qiu11,Jin09,Li10,Jakob10,liu15,Yin16,Wang16,Wang17}. Recently, various deep learning models~\cite{liu15,Yin16,Wang16,Wang17} have been developed and shown promising results for fine-grained sentiment analysis. However, most of existing methods under this branch only focus on terms extraction without categorizing them into specific categories. Though topic models or clustering based approaches~\cite{Ivan08,Lu09,Zhao10,chen14,Su08,Yu2011,Chen16} are able to group potential aspect terms into different clusters or topics (not explicit categories), they still fail to explicitly extract and classify a term into a predefined category.

For the task of aspect categorization, most existing methods assume the aspect terms be extracted in advance, and aim to predict their corresponding categories~\cite{Carenini05,Yu2011,Zhai10,Zhai11}. To apply these methods to in our problem setting, one needs to first identify aspect/opinion terms as a preprocessing step. In such a pipeline solution, error can be propagated across steps.

Multi-task learning aims to improve generalization for each individual task by exploiting relatedness among different tasks~\cite{Caruana97}. One common assumption in multi-task learning is that parameters for different tasks lie in a low-dimensional subspace~\cite{Argyriou08,Kumar12} which is achieved either by imposing low-rank constraint or matrix factorization. Through factorization, the model of each task becomes a linear combination of a small set of latent tasks. Following this idea, a multi-linear model was proposed in~\cite{Romera13} to deal with multi-modal tasks with multiple indexes. This tensor factorization idea also promotes a deep multi-task learning model~\cite{Yang16} where the parameters in different layers of a CNN for different tasks form a tensor that could be factorized across tasks. Moreover, many deep learning models have been introduced for multi-task learning~\cite{Liu15mtl,Ishan16} with an aim to learn shared hidden representation that are regularized from different tasks. Our proposed deep multi-task learning model is specially designed for sentiment analysis and is expected to be more effective for the proposed finer-grained sentiment analysis problem compared with other general methods.

\section{Problem Statement and Motivation}
Let $\mathcal{C} \!=\! \{1, 2, ..., C\}$ denote a predefined set of $C$ categories, where $c \!\in\! \mathcal{C}$ is an entity/attribute type, e.g., ``DRINK\#QUALITY'' in the restaurant domain. A review sentence $i$ is represented by a $d\times n_i$ matrix $\mathbf{X}_{i} \!=\! \{\mathbf{x}_{i1}, ..., \mathbf{x}_{in_{i}}\}$, where $\mathbf{x}_{ij}\in\R^{D}$ is a feature vector for the $j$-th token of the sentence. Moreover, for training, a set of corresponding labels for the sentence is represented by a vector $\mathbf{y}^c_i\in\R^{n_i}$, where $y^c_{ij}\in\{\mbox{BA}_{c}, \mbox{IA}_{c}, \mbox{BP}_{c}, \mbox{IP}_{c}, \mbox{O}_{c}\}$ is the label of the $j$-th token. Here, $\mbox{BA}_{c}$ and $\mbox{IA}_{c}$ refer to \textit{beginning of aspect} and \textit{inside of aspect}, respectively, of \textbf{category} $c$, similar for $\mbox{BP}_{c}$ and $\mbox{IP}_{c}$ for opinions, and $\mbox{O}_{c}$ refers to \textit{none of aspect or opinion terms} w.r.t. category $c$. In the following, we use $j$ to denote the index of a token in a sentence, and for simplifying notations,  we omit the sentence index $i$ if the context is clear.

To solve the above problem using existing aspect/opinion term extraction methods, one straightforward solution is to apply an extraction model to identify general aspect/opinion terms first, and then post-classify them into different categories using an additional classifier. However, this pipeline approach may suffer from error propagation from the extraction phrase to the classification phrase. An alternative solution is to train an extraction model for each category $c$ independently, and then combine the results of all the extraction models to generate final predictions. However, for each fine-grained category, aspect and opinion terms become extremely sparse for training, which makes it difficult to learn a precise model for each category if trained independently.

To address the above issues, we propose to model the problem in a multi-task learning manner, where aspect/opinion terms extraction for each category is considered as an individual task, and an end-to-end deep learning architecture is developed to jointly learn the tasks by exploiting their commonalities and similarities. To be specific, there are 4 main components in our proposed deep architecture: 1) {\bf Dual Propagation Memory Network} to co-extract aspect and opinion terms for each category, 2) {\bf Shared Tensor Decomposition} to model the commonalities of syntactic relations among different categories by sharing the tensor parameters, 3) {\bf Context-aware Multi-task Feature Learning}, which aims to jointly learn features among categories through constructing context-aware task similarity matrices, and 4) {\bf Auxiliary Task}, which creates an auxiliary task to predict overall sentence-level category labels to assist token-level prediction tasks. In the sequel, we name our proposed model as Multi-task Memory Networks (MTMNs). In the following section, we present the 4 components in detail.

\begin{figure}
	\centering
    \vspace{-4mm}
	\includegraphics[width=0.9\textwidth]{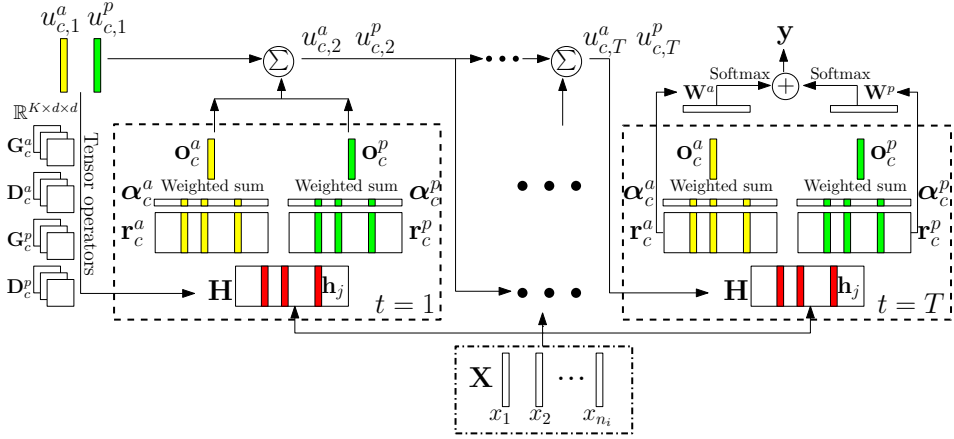}
    \vspace{-2mm}
	\caption{Architecture of dual propagation memory networks for aspect \& opinion terms extraction for category $c$.}\label{fig::dpMN}
    \vspace{-2mm}
	\label{fig::overall}
\end{figure}

\section{Proposed Deep Architecture}\label{sec::method}
\subsection{Dual Propagation Memory Network}\label{sec::modelSingle}
In this section, we present the base classifier used in MTMN for aspect and opinion terms co-extraction for each category $c$. The basic idea is to exploit syntactic relations to double propagate label information between aspects and opinions ~\cite{Qiu09} for extraction through a memory mechanism. Specifically, given a sentence with pre-trained word embeddings $\mathbf{X}\!=\![\mathbf{x}_1,...,\mathbf{x}_{n_i}]$, we first apply Gated Recurrent Unit (GRU)~\cite{Cho14} to obtain a memory matrix $\mathbf{H}\!=\![\mathbf{h}_{1}, ..., \mathbf{h}_{n_i}]$, where $\mathbf{h}_j\in\R^d$ is a feature vector. The architecture of the dual propagation memory network is shown in Figure~\ref{fig::dpMN}. At the first layer, an aspect prototype vector $\mathbf{u}_c^a$ and an opinion prototype vector $\mathbf{u}_c^p$ (denoted by $\mathbf{u}_{c,1}^a$ and $\mathbf{u}_{c,1}^p$ in the figure) are initialized in order to guide the attention to select aspect and opinion terms for category $c$. We then compute the interaction between $\mathbf{u}_c^p$, $\mathbf{u}_c^a$ and each memory $\mathbf{h}_j$ via tensor operations, respectively:
\begin{eqnarray}
\mathbf{r}_{c[j]}^{a}=\tanh([\mathbf{h}_j^{\top}\mathbf{G}_c^{a}\mathbf{u}_c^{a}:\mathbf{h}_j^{\top}\mathbf{D}_c^{a}\mathbf{u}_c^{p}]), \mbox{ and }
\mathbf{r}_{c[j]}^{p}=\tanh([\mathbf{h}_j^{\top}\mathbf{G}_c^{p}\mathbf{u}_c^{a}:\mathbf{h}_j^{\top}\mathbf{D}_c^{p}\mathbf{u}_c^{p}]), \label{eqn::r}
\end{eqnarray}
where $[:]$ denotes concatenation of vectors, the subscript $_{[j]}$ denotes the $j$-th column of the associated matrix, and $\mathbf{G}_c^{a},\mathbf{G}_c^{p},\mathbf{D}_c^{a},\mathbf{D}_c^{p}\in\R^{K\times d\times d}$ are 3-dimensional tensors composed of $K$ bi-linear interaction matrices. Intuitively, $\mathbf{G}_c^{a}$ or $\mathbf{D}_c^{p}$ is to capture the $K$ syntactic relations within aspect terms or opinion terms themselves, while $\mathbf{G}_c^{p}$ and $\mathbf{D}_c^{a}$ are to capture syntactic relations between aspect terms and opinion terms. Note that $\mathbf{r}_{c[j]}^{a}$ and $\mathbf{r}_{c[j]}^{p}$, both of which are of $2K$ dimensions, are the hidden representations for $\mathbf{h}_j$ w.r.t. aspect and opinion of category $c$, respectively\footnote{We further apply a GRU layer on top of (\ref{eqn::r}). In the subsequent computations, the notation $\mathbf{r}_{c}^{a}$ and $\mathbf{r}_{c}^{p}$ represent the features obtained after GRU.}.

With $\mathbf{r}_{c[j]}^{a}$ and $\mathbf{r}_{c[j]}^{p}$, normalized attention scores for $\mathbf{h}_j$ are computed as follows, which indicate the chance of the $j$-th token being an aspect and an opinion of category $c$. 
\begin{equation}
\boldsymbol{\alpha}^{a}_{c[j]} = \exp(\mathbf{e}^{a}_{c[j]})/\sum_{k}\exp(\mathbf{e}^{a}_{c[k]}), \mbox{ and } \boldsymbol{\alpha}^{p}_{c[j]} = \exp(\mathbf{e}^{p}_{c[j]})/\sum_{k}\exp(\mathbf{e}^{p}_{c[k]}),
\end{equation}
where $\boldsymbol{\alpha}^{a}_{c[j]}$ denotes the $j$-th element of the vector $\boldsymbol{\alpha}^{a}_{c}$, similar for $\mathbf{e}_{c[j]}$. Here $\mathbf{e}^{a}_{c[j]} \!=\! \langle{\mathbf{v}^{a}}, \mathbf{r}^{a}_{c[j]}\rangle$, and $\mathbf{e}^{p}_{c[j]} \!=\! \langle{\mathbf{v}^{p}}, \mathbf{r}^{p}_{c[j]}\rangle$. The overall representations of the sentence in terms of aspect and opinion are then computed as the weighted sums of $\{\mathbf{h_j}\}$'s using aspect and opinion attentions, respectively,
\begin{equation}
\mathbf{o}_c^{a}=\sum_{j}\boldsymbol{\alpha}^{a}_{c[j]}\mathbf{h}_{j}, \mbox{ and } \mathbf{o}_c^{p}=\sum_{j}\boldsymbol{\alpha}^{p}_{c[j]}\mathbf{h}_{j}.
\end{equation}
Intuitively, $\mathbf{o}_c^{a}$ and $\mathbf{o}_c^{p}$ are dominated by the input feature vectors $\{\mathbf{h}_i\}$'s with higher aspect and opinion attention scores, respectively, and will help to produce better prototype vectors for the next layer:
\begin{equation}\label{eq::prototypeUpdate}
\mathbf{u}^{a}_{c,t+1} = \tanh(\mathbf{Q}^{a} \mathbf{u}^{a}_{c,t})+\mathbf{o}^{a}_{t}, \mbox{ and } \mathbf{u}^{p}_{c,t+1} = \tanh(\mathbf{Q}^{p} \mathbf{u}^{p}_{c,t})+\mathbf{o}^{p}_{t},
\end{equation}
where the subscript $t$ is the index of a layer, and the matrices $\mathbf{Q}^{a}$ and $\mathbf{Q}^{p}$ are transformations. In this way, $\mathbf{u}^{a}_{c,t+1}$ (or $\mathbf{u}^{p}_{c,t+1}$) incorporates most probable aspect (or opinion) terms, which in turn will be used to interact with $\{\mathbf{h}_i\}$'s at layer $t\!+\!1$ to learn more precise token representations, attention scores, and sentence representations for selecting other non-obvious target tokens.

At the last layer $T$, after generating all the $\{\mathbf{r}^{a}_{c[j]}\}$'s and $\{\mathbf{r}^{p}_{c[j]}\}$'s, for each category $c$, we compute two 3-dimensional label vectors $\mathbf{y}^{a}_{c[j]}$ and $\mathbf{y}^{p}_{c[j]}$ as follows,
\begin{equation}
\mathbf{y}^{a}_{c[j]} = \mbox{softmax}(\mathbf{W}^a\mathbf{r}^{a}_{c[j]}), \mbox{ and } \mathbf{y}^{p}_{c[j]} = \mbox{softmax}(\mathbf{W}^p\mathbf{r}^{p}_{c[j]}),
\end{equation}
where $\mathbf{W}^a, \mathbf{W}^p\in\R^{3\times 2K}$ are transformation matrices for aspect and opinion prediction, respectively, and $\mathbf{y}^{a}_{c[j]}$ denotes the probabilities of $\mathbf{h}_j$ being BA$_c$, IA$_c$ and O$_c$, while $\mathbf{y}^{p}_{c[j]}$ denotes the probabilities of $\mathbf{h}_j$ being BP$_c$, IP$_c$ and O$_c$. For training, we define the loss function as follows,
\begin{equation}\label{eq::lossTok}
\mathcal{L}_{\mbox{tok}} = \sum_{c}\sum_{j=1}^{n_i}\sum_{m\in\{a,p\}}\ell\left({\boldsymbol{\hat{y}}^{m}_{c[j]}}, \mathbf{y}^{m}_{c[j]}\right),
\end{equation}
where $\ell(\cdot)$ is the cross-entropy loss, and $\boldsymbol{\hat{y}}^{m}_{c[j]}\in\R^{3}$ is a one-hot vector representing ground-truth label for $j$-th token w.r.t. aspect or opinion. For testing or making predictions, we generate a label for each token $j$ as follows. We first produce a label $\mathbf{y}_{c[j]}$ for category $c$ on the $j$-th token by comparing the largest value in $\mathbf{y}^{a}_{c[j]}$ and $\mathbf{y}^{p}_{c[i]}$. If both of them are O$_c$, then the label is O$_c$. If only one of them is O$_c$, we pick the other one as the label. Otherwise, the label is the one with the largest value. We then generate the final label on the $j$-th token by integrating $\mathbf{y}_{c[j]}$'s across all the categories, which is similar to multi-label classification, because some word might belong to multiple categories.

Note that a similar idea was proposed in~\cite{Wang17} for overall aspect and opinion terms extraction. However, as will be shown in experiments, applying their work directly to solve our proposed finer-grained extraction problem fails to achieve satisfactory performance. This is because in our proposed finer-grained extraction problem, training data for each specific category becomes too sparse to learn precise predictive models if extractions for different categories are considered independently. In the following sections, we introduce how to encode multi-task learning techniques to make the dual propagation memory network effective for finer-gained sentiment analysis.

\begin{figure}
	\centering
    \vspace{-4mm}
	\includegraphics[width=0.5\textwidth]{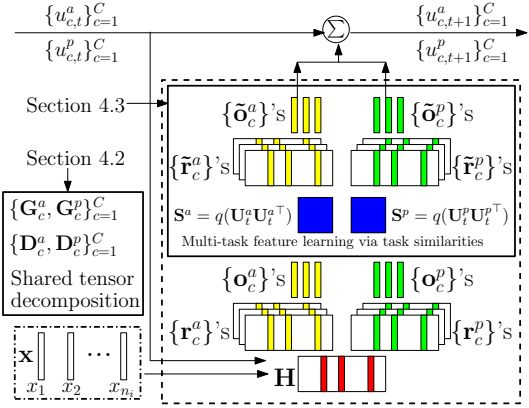}
    \vspace{-4mm}
    \vspace{1mm}
	\caption{The architecture of each non-output layer used in MTMNs.}
    \vspace{-2mm}
	\label{fig::MTMNnonOutput}
\end{figure}

\subsection{Shared Tensor Decomposition}
As described in the previous section, for each category $c$, there are four tensor operators $\mathbf{G}^a_c$, $\mathbf{G}^p_c$, $\mathbf{D}^a_c$, and $\mathbf{D}^p_c$, each of which is in $\R^{K\times d\times d}$, to model the complex token interactions. However, when the number of categories increases, the parameter size may be very large. As a result, available training data may be too sparse to learn the precise parameters. Therefore, instead of learning the tensors for each category independently, we assume that interactive relations among tokens are similar across categories. Therefore, we propose to learn a low-rank shared information among the tensors through collective tensor factorization as shown in the architecture of each non-output layer in Figure~\ref{fig::MTMNnonOutput}. Specifically, let $\mathbf{G}^a \!\in\! \R^{C\times K\times d\times d}$ be the concatenation of all the $\{\mathbf{G}^a_{c}\}$'s, and denote by $\mathbf{G}^a_k \!=\! {\mathbf{G}^a}_{[\cdot,k,\cdot,\cdot]} \!\in\! \R^{C\times d\times d}$ the collection of $k$-th bi-linear interaction matrices across $C$ tasks for aspect attention. The same also applies to $\mathbf{G}^p$ and $\mathbf{G}^p_k$ for opinion attention. Factorization is performed on each $\mathbf{G}^a_k$ and $\mathbf{G}^p_k$, respectively, via
\begin{equation}
{\mathbf{G}^a_k}_{[c,\cdot,\cdot]} = {\mathbf{Z}^a_k}_{[c,\cdot]}\boldsymbol{\mathcal{G}}^a_k, \mbox{ and } {\mathbf{G}^p_k}_{[c,\cdot,\cdot]} = {\mathbf{Z}^p_k}_{[c,\cdot]}\boldsymbol{\mathcal{G}}^p_k,
\end{equation}
where $\boldsymbol{\mathcal{G}}^a_k,\boldsymbol{\mathcal{G}}^p_k \!\in\! \R^{m\times d\times d}$ are shared factors among all the tasks with $m \!<\! C$, while $\mathbf{Z}^a_k,\mathbf{Z}^p_k \!\in\! \R^{C\times m}$ with each row ${\mathbf{Z}^a_k}_{[c,\cdot]}$ and ${\mathbf{Z}^p_k}_{[c,\cdot]}$ being specific factors for category $c$. The shared factors can be considered as $m$ latent basis interactions, where the original $k$-th bi-linear relation matrix ${\textbf{G}^a_k}_{[c,\cdot,\cdot]}$ (${\textbf{G}^p_k}_{[c,\cdot,\cdot]}$) for $c$ is the linear combination of the latent basis interactions. The same approach also applies to the tensors $\{\mathbf{D}^a_{c}\}$'s and $\{\mathbf{D}^p_{c}\}$'s. In this way, we reduce the parameter dimensions by enforcing sharing within a small number of latent interactions.

\subsection{Context-aware Multi-task Feature Learning}
Besides jointly decomposing tensors of syntactic relations across categories, in this section, we further exploit similarities between categories or tasks\footnote{We interchangeably use the terms ``task'' and ``category'' in the rest of this paper.} to learn more powerful features for each token and each sentence. Consider the following motivating example, ``FOOD\#PRICE'' is more similar to ``DRINK\#PRICE'' than ``SERVICE\#GENERAL'' because the first two categories may share some common aspect/opinion terms, such as \textit{expensive}. Therefore, by representing each task in a form of distributed vector, we can directly compute their similarities to facilitate knowledge sharing. Based on this motivation, we aim to update features $\boldsymbol{\tilde{r}}_c^{a}$ (or $\boldsymbol{\tilde{r}}_c^{p}$) from $\mathbf{r}_c^{a}$ (or $\mathbf{r}_c^{p}$) by integrating task relatedness. Specifically, at a layer $t$, suppose $\mathbf{u}^a_{c,t}$, and $\mathbf{u}^p_{c,t}$ are the updated prototype vectors passed from the previous layer, they can be used to represent task $c$. Because $\mathbf{u}^a_{c,t}$ and $\mathbf{u}^p_{c,t}$ are learned interactively with the category-specific sentence representations $\mathbf{o}^a_{c}$'s and $\mathbf{o}^p_{c}$'s of the previous $t\!-\!1$ layers, respectively. Let $\mathbf{U}^a$ and $\mathbf{U}^p\in\R^{d\times C}$ denote the matrices consisting of $\mathbf{u}^a_{c}$ and $\mathbf{u}^p_{c}$ as a column vector, respectively, then the task similarity matrices, $\mathbf{S}^a$ and $\mathbf{S}^p$, in terms of aspects and opinions can be computed as follows,
\begin{equation}
\mathbf{S}^a = q({\mathbf{U}^a}^\top{\mathbf{U}^a}), \mbox{ and } \mathbf{S}^p = q({\mathbf{U}^p}^\top{\mathbf{U}^p}),
\end{equation}
where $q(\cdot)$ is the softmax function carried in a column-wise manner so that the similarity scores between a task and all the tasks sum up to 1. The similarity matrices $\mathbf{S}^a$ and $\mathbf{S}^p$ are then used to refine feature representation of each token for each task by incorporating feature representations from related tasks:
\begin{equation}
\boldsymbol{\tilde{r}}^{a}_{c,[j]} = \sum_{c'=1}^{C}\mathbf{S}^a_{cc'} \mathbf{r}^{a}_{c',[j]}, \mbox{ and } \boldsymbol{\tilde{r}}^{p}_{c[j]} = \sum_{c'=1}^{C}\mathbf{S}^p_{cc'} \mathbf{r}^{p}_{c'[j]},
\end{equation}
where $\mathbf{r}^{a}_{c',[j]}$ and $\mathbf{r}^{p}_{c',[j]}$ denote the $j$-th column of the matrix $\mathbf{r}^{a}_{c'}$ and $\mathbf{r}^{p}_{c'}$, respectively. Similarly, we refine feature representation of each sentence for each task as follows,
\begin{eqnarray}
\boldsymbol{\tilde{o}}^a_{c} = \sum_{c'=1}^{C}\mathbf{S}^a_{cc'} \mathbf{o}^a_{c'}, \mbox{ and } \boldsymbol{\tilde{o}}^p_{c} = \sum_{c'=1}^{C}\mathbf{S}^p_{cc'} \mathbf{o}^p_{c'}.
\end{eqnarray}
Regarding update of the prototype vectors in (\ref{eq::prototypeUpdate}), we replace $\mathbf{o}^a_{c}$ and $\mathbf{o}^p_{c}$ by $\boldsymbol{\tilde{o}}^a_{c}$ and $\boldsymbol{\tilde{o}}^p_{c}$, respectively. This context-aware multi-task architecture is also shown in Figure~\ref{fig::MTMNnonOutput}. Note that this feature sharing among different tasks is context-aware because $\mathbf{U}^a$ and $\mathbf{U}^p$ are category representations depending on each sentence. This means that different sentences might indicate different task similarities. For example, when \textit{cheap} is presented, it might increase the similarity between ``FOOD\#PRICES'' and ``RESTAURANT\#PRICES''. As a result, $\boldsymbol{\tilde{r}}^{a}_{c[j]}$ for task $c$ could incorporate more information from task $c'$ if $c'$ has higher similarity score indicated by $\mathbf{S}^{a}_{cc'}$.

\subsection{Auxiliary Task}
As shown in previous sections, in MTMNs, besides learning token-level feature representations, sentence-level feature representations are learned as well. Thus, to better address the data sparsity issue, we aim to use additional global information on categories in the sentence level. Consider the following motivating example, if we know the sentence ``\textit{The soup is served with nice portion, the service is prompt}'' belongs to category ``DRINKS\#STYLE\_OPTIONS'' and ``SERVICE\#GENERAL'', we can infer that some words in the sentence should belong to one of these two categories. To make use of this information, we construct an auxiliary task to predict the categories of a sentence. From training data, sentence-level labels can be automatically obtained by integrating tokens' labels. Therefore, besides the token loss (\ref{eq::lossTok}) for our target token-level prediction tasks, we also define sentence loss for the auxiliary task. Note that the learning of the target task (terms extraction) and auxiliary task (multi-label classification on sentences) are not independent. On the one hand, the global sentence information helps the attentions to select category-relevant tokens. On the other hand, if the attentions are able to attend to target terms, the output context representation will filter out irrelevant noise, which helps the overall sentence prediction.
\begin{figure}
	\centering
	\includegraphics[width=0.5\textwidth]{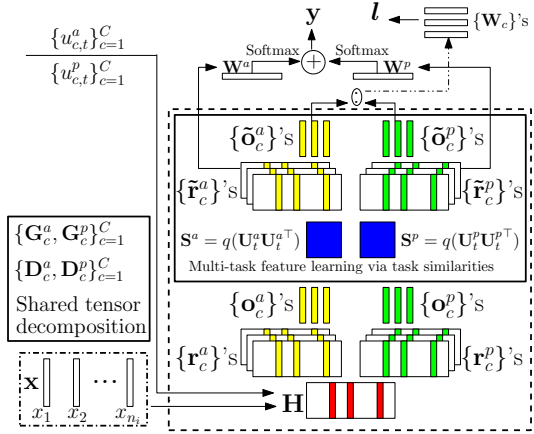}
    \vspace{-3mm}
	\caption{The architecture of the output layer used in MTMNs.}
	\label{fig::MTMNOutput}
\end{figure}
To be specific, as shown in Figure~\ref{fig::MTMNOutput}, for category $c$, we define $\boldsymbol{\tilde{o}}_c \!=\! [\boldsymbol{\tilde{o}}^a_{c}\!:\!\boldsymbol{\tilde{o}}^p_{c}] \!\in\! \R^{2d}$ the final representation for the sentence, and generate the output using the softmax function,
\begin{equation}
\boldsymbol{l}_c = \mbox{softmax}(\mathbf{W}_c\boldsymbol{\tilde{o}}_c),
\end{equation}
where $\mathbf{W}_c \!\in\! \R^{2\times 2d}$, and $\boldsymbol{l}_c \!\in\! \R^{2}$ indicates the probability of the sentence belonging to category $c$ or not. The loss of the auxiliary task is defined as $\mathcal{L}_{\mbox{sen}} = \sum_{c}\ell(\boldsymbol{\hat{l}}_c, \boldsymbol{l}_c)$, where $\ell(\cdot)$ is the cross-entropy loss, and $\boldsymbol{\hat{l}}_{c}\!\in\!\{0,1\}^2$ is the ground truth using one-hot encoding indicating whether category $c$ is presented for the sentence. By incorporating the loss of the auxiliary task, the final objective for MTMNs is written as $\mathcal{L} \!=\! \mathcal{L}_{\mbox{sen}} \!+\! \mathcal{L}_{\mbox{tok}}$, where $\mathcal{L}_{\mbox{tok}}$ is defined in (\ref{eq::lossTok}).

\section{Experiments}
\begin{wraptable}{r}{7cm}
\scriptsize
	\begin{center}\vspace{-5mm}
		\begin{tabular}{l|c|c|c|c|c|c}
			\hline
			\hline \multirow{2}{*}{Dataset} & \multicolumn{2}{c|}{Training} & \multicolumn{2}{c|}{Test} & \multicolumn{2}{c}{Total} \\
            \cline{2-7} & text & tuple & text & tuple & text & tuple \\
			\hline S1 (15) & 1,315 & 1,654 & 685 & 845 & 2,000 & 2,499 \\
            \hline S2 (16) & 2,000 & 2,507 & 676 & 859 & 2,676 & 3,366 \\
            \hline S3 (14) & 2,853 & 1,974 & 742 & 545 & 3,595 & 2,519 \\
			\hline
		\end{tabular}\vspace{-1mm}
		\caption{Dataset description.}\label{table::data}
	\end{center}\vspace{-4mm}
\end{wraptable}
The experiments are conducted on three benchmark datasets from subtask 1 in SemEval Challenge 2015 task 12~\cite{sem15}, SemEval Challenge 2016 task 5~\cite{Sem16}, and SemEval Challenge 2014 task 4~\cite{sem14}, which are denoted by S1, S2, and S3, respectively. Note that S1 and S2 are reviews in restaurant domain and S3 is in laptop domain. We use term-level aspect-opinion annotations provided by~\cite{Wang17} for S1 and S3, and manually annotate opinion terms for S2. To facilitate our experiment, we additionally annotate category labels on target terms for S3, while the aspect term categories for S1 and S2 are provided by SemEval. The statistics of each dataset is shown in Table~\ref{table::data} where \textit{text} and \textit{tuple} represent the number of sentences and the number of tuples consisting of an aspect term and its corresponding category label, respectively. Each sentence may contain multiple aspect terms with more than one categories. The aspect categories are shown in Table~\ref{table::category}.\footnote{We filter out some categories with very few target terms.} For S1 and S2, an aspect category is defined as the combination of an entity and an attribute, e.g., ``FOOD\#PRICES''. There are in total 12 categories. For S3, an aspect category is an entity.

Follow~\cite{Wang17}, we first obtain word embeddings by applying \textit{word2vec}\footnote{https://radimrehurek.com/gensim/models/word2vec.html} on Yelp Challenge dataset\footnote{http://www.yelp.com/dataset\_challenge} and electronic domain in Amazon reviews~\cite{Mc15} for restaurant and laptop datasets, respectively. We set the dimension of word embeddings to be 150 and the dimension after GRU transformation to be 50 for all the three datasets. We use two layers as in the proposed MTMNs for experiments. For each layer, the number of bi-linear interactions for the 3-dimensional tensors is 20 ($K\!\!=\!\!20$), and tensor factorization operates with $m\!=\!5$ for S1 and S2, and $m\!=\!8$ for S3. We apply partial dropout at 0.5 to chosen parameters to avoid overfitting. The training is carried with \textit{rmsprop} with the initial value at 0.001 and decayed with rate 0.9. The trade-off parameter $\lambda$ is set to be 1.0. All the hyper-parameters are chosen according to cross-validation.
\begin{table}[h]
\begin{center}\vspace{-2mm}
\begin{adjustbox}{max width=\textwidth}
		\begin{tabular}{c|c|c}
        	\hline
            \hline
            \multicolumn{2}{c|}{\textbf{Restaurant}} & \multicolumn{1}{c}{\textbf{Laptop}} \\
			\hline
            Entity Labels & Attribute Labels & Entity Labels \\
			\hline
            1. RESTAURANT 2. FOOD 3. DRINKS    & A. GENERAL B. PRICES C. QUALITY    & 1. LAPTOP 2. DISPLAY 3. KEYBOARD 4. MOUSE 5. BATTERY  \\
            4. AMBIENCE 5. SERVICE 6. LOCATION & D. STYLE\_OPTIONS E. MISCELLANEOUS & 6. GRAPHICS 7. HARD\_DISC 8. MULTIMEDIA\_DEVICES \\
                                               &                                    & 9. SOFTWARE 10. OS 11. SUPPORT 12. COMPANY \\
            \hline
		\end{tabular}
\end{adjustbox}
		\vspace{1mm}
        \caption{Aspect Categories for two domains.}\label{table::category}
\end{center}\vspace{-10mm}
\end{table}

\subsection{Experimental Results}
We conduct comparison experiments with the following baseline models:

\noindent\textbf{NLANG}: The best system for both SemEval-15 and SemEval-16 for the proposed task.

\noindent\textbf{IHS-RD, XRCE}: The second best systems for SemEval-15 and SemEval-16, respectively.

\noindent\textbf{RNCRF+}: We modify RNCRF~\cite{Wang16}, which is for aspect-opinion terms extraction, by defining finer-grained categories as labels. This means directly apply RNCRF to a multi-class classification problem with $C\times 5$ classes, as there are 5 classes for each category and there are $C$ categories. 

\noindent\textbf{CMLA+}: Similar to RNCRF+, we modify CMLA~\cite{Wang17} by defining finer-grained categories as labels. Note that the proposed MTMNs can be reduced to this baseline model by only using a single dual propagation memory introduced in Section~\ref{sec::modelSingle} with $C\times 5$ classes.

\noindent\textbf{CMLA++}: CMLA is used to extract all the aspect and opinion terms first, and then a category classification layer is added to classify the extracted terms.

We report the results from top performing systems in the Challenges for S1 and S2. There are no reported results for S3 as the original task is different from ours. Note that original task for S1 and S2 includes two slots: slot 1 for sentence-level aspect category prediction and slot 2 for aspect terms extraction. Moreover, SemEval also evaluated on the pairing of slot 1 and slot 2 by joining them as an additional task that corresponds to the problem we study. However, most of the reported models did not provide feasible methods for the joint prediction of aspect terms and corresponding categories. Instead, they trained the model for slot 2 first and then combined with slot 1. This may fail to capture the relations between target terms and their categories. In order to show the advantage of our model, we modify the existing state-of-the-art deep models for aspect/opinion term extraction to fit our problem settings. Since RNCRF and CMLA both exploit the correlations between aspect terms and opinion terms, which have been shown to be effective for extraction task, a simple idea is to increase the number of classes to incorporate different categories, e.g., BA becomes $\{\mbox{BA}_c\}$'s for different category $c$. By increasing the number of classes, the only change to the original model is the dimension of classification matrix. As a result, the modified model should be able to capture the correlations between target terms based on their categories. On the other hand, we also construct another baseline model (denoted by CMLA++) based on CMLA by separating the task into 2 steps. The first step is the same as CMLA for extracting target terms. Then the second step performs category prediction only on the extracted terms.
\begin{table*}
\scriptsize
	\begin{center}
	\begin{adjustbox}{max width=\textwidth}
		\begin{tabular}{l|c|c|c|c|c|c|c|c|c|c|c|c}
			\hline
			\hline & \multicolumn{4}{c|}{S1} & \multicolumn{4}{c|}{S2} & \multicolumn{4}{c}{S3} \\
			\hline Model & \textbf{ASC} & \textbf{OPC} & AS & OP & \textbf{ASC} & \textbf{OPC} & AS & OP & \textbf{ASC} & \textbf{OPC} & AS & OP \\
            \hline NLANG & 42.90 & - & 67.11 & - & 52.61 & - & 72.34 & - & - & - & - & - \\
			\hline IHS\_RD & 42.72 & - & 63.12 & - & - & - & - & - & - & - & - & -\\
			\hline XRCE & - & - & - & - & 48.89 & - & 61.98 & - & - & - & - & - \\
			\hline RNCRF+ & 54.00 & 47.86 & 67.74 & 67.62 & 56.04 & 51.09 & 69.74 & 74.26 & 54.05 & 58.90 & 71.87 & 76.62 \\
            \hline CMLA+ & 57.35 & 55.70 & 70.73 & 73.68 & 57.83 & 56.04 & 75.21 & 77.90 & 55.71 & 62.40 & 72.42 & 76.98 \\
            \hline CMLA++ & 53.46 & 53.94 & 70.73 & 73.68 & 54.05 & 54.34 &75.21 & 77.90 & 54.31 & 62.95 & 72.42 & 76.98 \\
            \hline \textbf{MTMN} & \textbf{63.16} & \textbf{59.17} & \textbf{71.31} & 72.23 & \textbf{65.34} & \textbf{61.44} & 73.26 & 76.10 & \textbf{57.06} & \textbf{63.53} & 69.14 & 75.76\\
            \hline
		\end{tabular}\vspace{-2mm}
		\end{adjustbox}
		\caption{Comparison results in terms of $F_1$ scores. ASC (OPC) refers to category-specific aspect (opinion) terms extraction. AS (OS) refers to aspect (opinion) terms extraction.}\label{table:comparison}
		\vspace{-5mm}
	\end{center}
\end{table*}

The comparison results are shown in Table~\ref{table:comparison}. It can be seen that MTMN achieves the state-of-the-art performances in category-specific aspect and opinion terms extraction (ASC and OPC). And there is a large gap between the results of MTMN and the other baseline models on S1 and S2. This is because RNCRF+ and CMLA+ can only propagate information between target terms within each category, but fail to explore the relations and commonalities among different categories. The other model CMLA++ performs even poorer, because the training is separated into different stages, similar to the top systems in SemEval Challenges. This separation results in the failure of propagating information from category prediction to target term extraction. The result proves the effectiveness of MTMN for learning shared information among different tasks, as well as the addition of global information to assist extraction. The improvement for S3 is not significant, which might indicate that the category correlations are not obvious in laptop domain, as can be seen in Table~\ref{table::category}. Only entity labels make different categories distinct from each other.

Moreover, we also report the results on target terms extraction (AS and OP) by accumulating the aspect/opinion terms that are assigned at least one category by MTMN. It can be seen that MTMN still achieves comparable performances even if the data becomes sparser when adding the category information. On the contrary, the results for RNCRF+, CMLA+ and CMLA++ are obtained using the original models that ignore category labels.

As have been discussed in the previous sections, the multi-task memory network explores the commonalities and relations among tasks through both tensor sharing and feature sharing, as well as enhances prediction results by incorporating auxiliary labels. To test the effect of each component, we conduct comparison experiments for different combinations of these components as shown in Table~\ref{tab::setting}, where C1, C2 and C3 represents separate component for multi-task tensor sharing, context-aware feature sharing and auxiliary task, respectively. 
\begin{wraptable}{r}{7.0cm}
\scriptsize\vspace{-4mm}
	\begin{center}
		\begin{tabular}{l|c|c|c|c}
			\hline
			\hline Different & \multicolumn{2}{c|}{S1} & \multicolumn{2}{c}{S2} \\
			\cline{2-5} Components & ASC & OPC & ASC & OPC \\
            \hline MTMN (C1+C2+C3) & \textbf{63.16} & \textbf{59.17} & \textbf{65.34} & \textbf{61.44} \\
			C1+C3 & 61.95 & 58.57 & 63.30 & 59.16 \\
			C2+C3 & 61.67 & 55.89 & 60.86 & 58.68 \\
			C2+C3* & 61.30 & 55.30 & 62.68 & 58.93 \\
            C1+C2 & 60.67 & 56.97 & 61.29 & 58.16 \\
            C3 & 60.18 & 57.03 & 60.57 & 57.36 \\
            \hline
		\end{tabular}\vspace{-2mm}
		\caption{Comparison results with reductions.}\label{tab::setting}
		\vspace{-3mm}
	\end{center}
\end{wraptable}
Note that for C2+C3*, we use the same tensor across all the tasks. Clearly, the inclusion of either feature sharing (C2+C3) or tensor sharing (C1+C3) improves the results for most of the time compared to independent training (C3). Furthermore, tensor sharing is more beneficial than feature sharing. This shows that the commonalities in terms of token interactions are more obvious for different tasks. Moreover, either independent tensors (C2+C3) or the same tensor (C2+C3*) across tasks does not perform well. This indicates that it is still crucial to explore both the uniqueness and commonality of all the tasks, which are preserved in our proposed model. By comparing the results between C1+C2+C3 and C1+C2, we can see how auxiliary task contributes to the final prediction (2.49\% and 4.05\% increase for ASC in S1 and S2 respectively). This shows that the global information in the sentence level could enforce the correct predictions of each token within the sentence by reiterating the category information.

\begin{figure}
	\vspace{-6mm}\centering
	\includegraphics[width=0.50\textwidth]{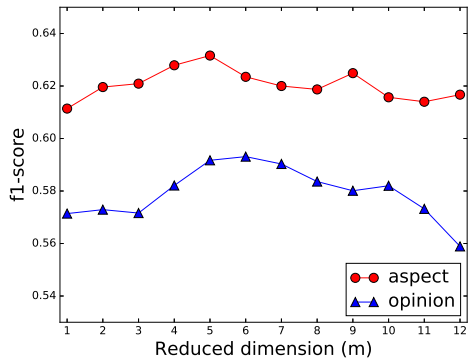}\vspace{-2mm}
	\caption{Sensitivity on S1.}
	\label{fig::sensitivity}
\end{figure}
To show the robustness of our model, we test it with different dimensions of factorization ($m$). The results on S1 are shown in Figure~\ref{fig::sensitivity}. We also provide some examples in Table~\ref{table::example} to show what the attentions learn for different categories. The second and third columns show the extracted aspect and opinion terms with corresponding normalized attention scores, respectively. The \textbf{bold numbers} denote category indexes. Clearly, MTMN is able to attend to relevant tokens for different categories. Moreover, MTMN could identify the case when specific terms belong to more than one categories.

\begin{table}[h]
\begin{center}\vspace{-2mm}
\begin{adjustbox}{max width=\textwidth}
		\begin{tabular}{c|c|c}
        	\hline
            \hline
            \multicolumn{1}{c|}{\textbf{Sentence}} & \multicolumn{1}{c|}{\textbf{Aspect (score)}} & \multicolumn{1}{c}{\textbf{Opinion (score)}} \\
			\hline
            \textit{Excellent} \textbf{food} though the \textbf{interior} could use some help. & \textbf{1}: food (0.77); \textbf{2}: interior (0.79) & \textbf{1}: Excellent (0.56) \\
			\hline
			The \textbf{sashimi} is always \textit{fresh} and the \textbf{rolls} are & \textbf{1}: sashimi (0.41), rolls (0.51) & \textbf{1}: fresh (0.52) \\
			\textit{innovative} and delicious. & \textbf{3}: rolls (0.49) & \textbf{3}: innovative (0.38) \\
			\hline
			The \textbf{food} was \textit{good}, the \textbf{place} was \textit{clean} and & \textbf{1}: food (0.70) & \textbf{1}: good (0.66); \textbf{2}: clean (0.46) \\
			\textit{affordable}. & \textbf{2}: place (0.81) & \textbf{4}: affordable (0.34) \\
			\hline
			Overall, \textit{decent} \textbf{food} at a \textit{good} price, with & \textbf{1}: food (0.70) & \textbf{1}: decent (0.31); \textbf{5}: good (0.28) \\
			\textit{friendly} \textbf{people}. & \textbf{6}: people (0.53) & \textbf{6}: friendly (0.38) \\
			\hline
			The \textbf{wine list} was \textit{expensive}, though the \textbf{staff} & \textbf{6}: staff (0.51) & \textbf{6}: knowledgeable (0.49) \\
			are not \textit{knowledgeable}. & \textbf{7}: wine (0.32) list (0.31) & \textbf{7}: extensive (0.64) \\
			\hline The \textbf{martinis} are \textit{amazing} and very \textit{fairly priced}. & \textbf{8}: martinis (0.42) & \textbf{8}: fairly (0.21) priced (0.69) \\
			& \textbf{9}: martinis (0.53) & \textbf{9}: amazing (0.36) \\
			\hline 
			\textit{Abulous} \textbf{food}, if the \textbf{front of house staff} do n't & \textbf{1}: food (0.82); \textbf{6}: front (0.26) &  \\
			put you off. & of (0.10) house (0.23) staff (0.21) & \textbf{1}: Abulous (0.46) \\
			\hline
			The \textbf{food} was \textit{great} and \textit{tasty}, but the \textbf{sitting space} & \textbf{1}: food (0.59) & \textbf{1}: great (0.36), tasty (0.35) \\
			was too \textit{small}. & \textbf{2}: sitting (0.35) space (0.27) & \textbf{2}: small (0.32) \\
			\hline
			I have never been so \textit{disgusted} by \textbf{food} and \textbf{service}. & \textbf{1}: food (0.86); \textbf{6}: service (0.66) & \textbf{1, 6}: disgusted (0.50, 0.76) \\
			\hline
			Despite the \textit{confusing} \textbf{mirrors} this will be my & \textbf{1}: Japanese (0.32) food (0.30) & \textbf{1}: go-to (0.50) \\
			\textit{go-to} for \textbf{Japanese food}. & \textbf{2}: mirrors (0.35) & \textbf{2}: confusing (0.39) \\
			\hline
			\textbf{Service} \textit{ok}, but \textit{unfriendly filthy} \textbf{bathroom}. & \textbf{2}: bathroom (0.79) & \textbf{2}: unfriendly (0.33), filthy (0.41) \\
			& \textbf{6}: service (0.52) & \textbf{6}: ok (0.31) \\

            \hline
		\end{tabular}
\end{adjustbox}
		\vspace{1mm}
        \caption{Examples of attention scores for different categories.}\label{table::example}
\end{center}\vspace{-10mm}
\end{table}

\section{Conclusion}
In this work, we introduce a finer-grained task involving the predictions of both aspect/opinion terms and their corresponding aspect categories, and offer a novel multi-task deep learning model, MTMN, to solve the problem. The model is able to exploit syntactic commonalities and task similarities through attention mechanism. In the end, we demonstrate the effectiveness of our model on three benchmark datasets.

\bibliographystyle{unsrt}
\bibliography{nips17}

\begin{thebibliography}{10}

\bibitem{Hu04}
Minqing Hu and Bing Liu.
\newblock Mining and summarizing customer reviews.
\newblock In {\em KDD}, pages 168--177, 2004.

\bibitem{Qiu11}
Guang Qiu, Bing Liu, Jiajun Bu, and Chun Chen.
\newblock Opinion word expansion and target extraction through double
  propagation.
\newblock {\em Comput. Linguist.}, 37(1):9--27, 2011.

\bibitem{Wang16}
Wenya Wang, Sinno~Jialin Pan, Daniel Dahlmeier, and Xiaokui Xiao.
\newblock Recursive neural conditional random fields for aspect-based sentiment
  analysis.
\newblock In {\em EMNLP}, 2016.

\bibitem{Carenini05}
Giuseppe Carenini, Raymond~T. Ng, and Ed~Zwart.
\newblock K-cap.
\newblock pages 11--18, 2005.

\bibitem{Yu2011}
Jianxing Yu, Zheng-Jun Zha, Meng Wang, and Tat-Seng Chua.
\newblock Aspect ranking: Identifying important product aspects from online
  consumer reviews.
\newblock In {\em ACL}, 2011.

\bibitem{Zhai10}
Zhongwu Zhai, Bing Liu, Hua Xu, and Peifa Jia.
\newblock Grouping product features using semi-supervised learning with
  soft-constraints.
\newblock In {\em COLING}, pages 1272--1280, 2010.

\bibitem{Zhai11}
Zhongwu Zhai, Bing Liu, Hua Xu, and Peifa Jia.
\newblock Clustering product features for opinion mining.
\newblock In {\em WSDM}, pages 347--354, 2011.

\bibitem{Guo09}
Honglei Guo, Huijia Zhu, Zhili Guo, XiaoXun Zhang, and Zhong Su.
\newblock Product feature categorization with multilevel latent semantic
  association.
\newblock In {\em CIKM}, pages 1087--1096, 2009.

\bibitem{Titov08}
Ivan Titov and Ryan McDonald.
\newblock Modeling online reviews with multi-grain topic models.
\newblock In {\em www}, pages 111--120, 2008.

\bibitem{Pop05}
Ana-Maria Popescu and Oren Etzioni.
\newblock Extracting product features and opinions from reviews.
\newblock In {\em EMNLP}, pages 339--346, 2005.

\bibitem{Zhuang06}
Li~Zhuang, Feng Jing, and Xiao-Yan Zhu.
\newblock Movie review mining and summarization.
\newblock In {\em CIKM}, pages 43--50, 2006.

\bibitem{Wu09}
Yuanbin Wu, Qi~Zhang, Xuanjing Huang, and Lide Wu.
\newblock Phrase dependency parsing for opinion mining.
\newblock In {\em EMNLP}, pages 1533--1541, 2009.

\bibitem{Jin09}
Wei Jin and Hung~Hay Ho.
\newblock A novel lexicalized hmm-based learning framework for web opinion
  mining.
\newblock In {\em ICML}, pages 465--472, 2009.

\bibitem{Li10}
Fangtao Li, Chao Han, Minlie Huang, Xiaoyan Zhu, Ying-Ju Xia, Shu Zhang, and
  Hao Yu.
\newblock Structure-aware review mining and summarization.
\newblock In {\em COLING}, pages 653--661, 2010.

\bibitem{Jakob10}
Niklas Jakob and Iryna Gurevych.
\newblock Extracting opinion targets in a single- and cross-domain setting with
  conditional random fields.
\newblock In {\em EMNLP}, pages 1035--1045, 2010.

\bibitem{liu15}
Pengfei Liu, Shafiq Joty, and Helen Meng.
\newblock Fine-grained opinion mining with recurrent neural networks and word
  embeddings.
\newblock In {\em EMNLP}, pages 1433--1443, 2015.

\bibitem{Yin16}
Yichun Yin, Furu Wei, Li~Dong, Kaimeng Xu, Ming Zhang, and Ming Zhou.
\newblock Unsupervised word and dependency path embeddings for aspect term
  extraction.
\newblock In {\em IJCAI}, 2016.

\bibitem{Wang17}
Wenya Wang, Sinno~Jialin Pan, Daniel Dahlmeier, and Xiaokui Xiao.
\newblock Coupled multi-layer tensor network for co-extraction of aspect and
  opinion terms.
\newblock In {\em AAAI}, 2017.

\bibitem{Ivan08}
Ivan Titov and Ryan~T. McDonald.
\newblock A joint model of text and aspect ratings for sentiment summarization.
\newblock In {\em ACL}, pages 308--316, 2008.

\bibitem{Lu09}
Yue Lu, ChengXiang Zhai, and Neel Sundaresan.
\newblock Rated aspect summarization of short comments.
\newblock In {\em WWW}, pages 131--140, 2009.

\bibitem{Zhao10}
Wayne~Xin Zhao, Jing Jiang, Hongfei Yan, and Xiaoming Li.
\newblock Jointly modeling aspects and opinions with a maxent-lda hybrid.
\newblock In {\em EMNLP}, pages 56--65, 2010.

\bibitem{chen14}
Zhiyuan Chen, Arjun Mukherjee, and Bing Liu.
\newblock Aspect extraction with automated prior knowledge learning.
\newblock In {\em ACL}, pages 347--358, 2014.

\bibitem{Su08}
Qi~Su, Xinying Xu, Honglei Guo, Zhili Guo, Xian Wu, Xiaoxun Zhang, Bin Swen,
  and Zhong Su.
\newblock Hidden sentiment association in chinese web opinion mining.
\newblock In {\em WWW}, pages 959--968, 2008.

\bibitem{Chen16}
Lu~Chen, Justin Martineau, Doreen Cheng, and Amit~P. Sheth.
\newblock Clustering for simultaneous extraction of aspects and features from
  reviews.
\newblock In {\em {NAACL}-{HLT}}, pages 789--799, 2016.

\bibitem{Caruana97}
Rich Caruana.
\newblock Multitask learning.
\newblock {\em Mach. Learn.}, 28(1):41--75, 1997.

\bibitem{Argyriou08}
Andreas Argyriou, Theodoros Evgeniou, and Massimiliano Pontil.
\newblock Convex multi-task feature learning.
\newblock {\em Mach. Learn.}, 73(3):243--272, 2008.

\bibitem{Kumar12}
Abhishek Kumar and Hal~Daumé III.
\newblock Learning task grouping and overlap in multi-task learning.
\newblock In {\em ICML}, 2012.

\bibitem{Romera13}
Bernardino Romera-Paredes, Hane Aung, Nadia Bianchi-Berthouze, and Massimiliano
  Pontil.
\newblock Multilinear multitask learning.
\newblock In {\em ICML (3)}, volume~28 of {\em JMLR Workshop and Conference
  Proceedings}, pages 1444--1452, 2013.

\bibitem{Yang16}
Zichao Yang, Diyi Yang, Chris Dyer, Xiaodong He, Alex Smola, and Eduard Hovy.
\newblock Hierarchical attention networks for document classification.
\newblock In {\em NAACL}, pages 1480--1489, 2016.

\bibitem{Liu15mtl}
Xiaodong Liu, Jianfeng Gao, Xiaodong He, Li~Deng, Kevin Duh, and Ye-Yi Wang.
\newblock Representation learning using multi-task deep neural networks for
  semantic classification and information retrieval.
\newblock In {\em NAACL}, pages 912--921, 2015.

\bibitem{Ishan16}
Ishan Misra, Abhinav Shrivastava, Abhinav Gupta, and Martial Hebert.
\newblock Cross-stitch networks for multi-task learning.
\newblock 2016.

\bibitem{Qiu09}
Guang Qiu, Bing Liu, Jiajun Bu, and Chun Chen.
\newblock Expanding domain sentiment lexicon through double propagation.
\newblock In {\em IJCAI}, pages 1199--1204, 2009.

\bibitem{Cho14}
Kyunghyun Cho, Bart van Merrienboer, Caglar Gulcehre, Dzmitry Bahdanau, Fethi
  Bougares, Holger Schwenk, and Yoshua Bengio.
\newblock Learning phrase representations using rnn encoder-decoder for
  statistical machine translation.
\newblock In {\em EMNLP}, pages 1724--1734, 2014.

\bibitem{sem15}
Maria Pontiki, Dimitris Galanis, Haris Papageorgiou, Suresh Manandhar, and Ion
  Androutsopoulos.
\newblock {SemEval}-2015 task 12: Aspect based sentiment analysis.
\newblock In {\em SemEval 2015}, pages 486--495, 2015.

\bibitem{Sem16}
Maria Pontiki, Dimitrios Galanis, Haris Papageorgiou, Ion Androutsopoulos,
  Suresh Manandhar, Mohammad AL-Smadi, Mahmoud Al-Ayyoub, Yanyan Zhao, Bing
  Qin, Orphée~De Clercq, Véronique Hoste, Marianna Apidianaki, Xavier
  Tannier, Natalia Loukachevitch, Evgeny Kotelnikov, Nuria Bel, Salud~María
  Jiménez-Zafra, and Gülşen Eryiğit.
\newblock {SemEval}-2016 task 5: Aspect based sentiment analysis.
\newblock In {\em SemEval 2016}, 2016.

\bibitem{sem14}
Maria Pontiki, Dimitris Galanis, John Pavlopoulos, Harris Papageorgiou, Ion
  Androutsopoulos, and Suresh Manandhar.
\newblock Semeval-2014 task 4: Aspect based sentiment analysis.
\newblock In {\em SemEval}, pages 27--35, 2014.

\bibitem{Mc15}
Julian McAuley, Christopher Targett, Qinfeng Shi, and Anton van~den Hengel.
\newblock Image-based recommendations on styles and substitutes.
\newblock In {\em SIGIR}, pages 43--52, 2015.

\end{thebibliography}


\end{document}